\documentclass[letterpaper, 10 pt, conference]{ieeeconf}  

\IEEEoverridecommandlockouts                              

\usepackage{cite}
\usepackage{amsmath,amssymb,amsfonts}
\usepackage{algorithmic}
\usepackage{graphicx}
\usepackage{textcomp}
\usepackage{xcolor}
\usepackage[letterpaper, top=0.79in, bottom=0.77in, left=0.75in, right=0.75in, includehead=false, includefoot=false]{geometry}

\usepackage{accents}
\usepackage{algorithm}
\usepackage{caption}

\usepackage[absolute,overlay]{textpos}

\title{\LARGE \bf
A Switching Nonlinear Model Predictive Control Strategy for Safe Collision Handling by an Underwater Vehicle-Manipulator System
}

\author{Ioannis G. Polyzos$^{1}$ and Konstantinos J. Kyriakopoulos$^{2}$, Fellow, IEEE
\thanks{$^{1}$Ioannis G. Polyzos is with the D. Guggenheim School of Aerospace Engineering, Georgia Institute of Technology, Atlanta, GA, USA, 
        {\tt\small ipolyzos3@gatech.edu}}%
\thanks{$^{2}$Konstantinos J. Kyriakopoulos is with the faculty of Electrical Engineering, Engineering Division, New York University Abu Dhabi, Abu Dhabi, UAE,
        {\tt\small kkyria@nyu.edu}}%
}

\begin{document}

\maketitle
\thispagestyle{empty}
\pagestyle{empty}

\begin{textblock*}{\textwidth}(1.1in, 0.3in) 
    \noindent\footnotesize\textcolor{gray}{This work has been submitted to the 2026 Mediterranean Conference on Control and Automation (MED) to be considered for publication.}
\end{textblock*}

\begin{abstract}
For active intervention tasks in underwater environments, the use of autonomous vehicles is just now emerging as an active area of research. During operation, for various reasons, the robot might find itself on a collision course with an obstacle in its environment. In this paper, a switching Nonlinear Model Predictive Control (NMPC) strategy is proposed to safely handle collisions for an Underwater Vehicle-Manipulator System (UVMS). When avoiding the collision is impossible, the control algorithm takes advantage of the manipulator, using it to push against the obstacle, and deflect away from the collision. Virtual experiments are performed to demonstrate the algorithm's capability to successfully detect collisions and either avoid them, or use the manipulator to handle them appropriately without damaging sensitive areas of the vehicle.
\end{abstract}

\section{INTRODUCTION}

Over the last decades, interest has developed around marine applications, including underwater tasks that are categorized into survey and intervention missions. Survey missions are observational and focus on data collection using sensors, like cameras and sonar. Examples of such tasks are seafloor mapping, pipeline and cable inspections, coral reef monitoring, or search and rescue missions. Intervention tasks, additionally require physical interaction with objects found in the underwater environment of the robot. Their goal is to perform some sort of action through the End-Effector (EE) of a manipulator and/or some other tool. Underwater manipulators are tasked with salvage of sunken objects, cleaning surfaces, opening and closing valves, drilling, rope cutting, cable laying and repair, clearing debris and fishing nets, biological and geological sampling, among others \cite{UVMS_Survey_19}. These tasks are significantly more complex and require precise control and decision-making with the use of a remotely-operated UVMS, operated either from a mothership or from land, remaining the norm to this day. 

\subsection{Related Works}
Since the deployment of Autonomous Underwater Vehicles (AUV) has been mainly focused on survey missions, position tracking strategies via PID control are quite common \cite{PITracking}. Barbalata et al. \cite{AdaptivePID} also suggest the use of an adaptive PID-like position-velocity controller. More recently, there have been a series of publications on the use of MPC for the robust control of an UVMS. Dai et al. \cite{Dai_1} \cite{Dai_2} \cite{Dai_3} use an extended Kalman Filter with a Fast Tube MPC for trajectory and target tracking. Nikou et al. \cite{Nikoutube} extend the concept of tube MPC to UVMS interaction control. They also explore the concept of controlling a UVMS by providing high-level tasks \cite{NikouHighLevelTasks}. Heshmati-Alamdari et al. \cite{ShahabForceControl} \cite{ShahabInteraction} focus their work on model-free force and position control for UVMS interaction missions and provide performance guarantees.

For robot operation in environments where obstacles are present, decades of research has been focused on collision avoidance \cite{DynamicWindowApproach}, primarily achieved by path planning algorithms \cite{PotentialFieldMethods} \cite{MobileRobotPathPlanningReview}. A key drawback of these methods is the fact they provide a trajectory with uncertain feasibility when it comes to the robot dynamics. In search of safety guarantees, recent advances in control theory have led to the use of control barrier functions or Lyapunov functions \cite{Ames_CBF_Surevy}. While these methods offer an effective solution to the safe control problem under nominal circumstances, there exist emergency scenarios and unforeseen circumstances where they could possibly fail due to a force majeure event. Wosch et al. \cite{CollisionHandling_MobileManipulator} considered that emergency scenarios might arise for a terrestrial mobile manipulator, thus implementing a collision handling algorithm into their collision avoidance strategy. However their work does not predict collisions, but only detects and deals with them - to mitigate further unwanted effects - after they occur, posing significant questions regarding robot safety. In \cite{MartialArts}, Chen et al. deal with emergency scenarios stemming from fast moving unexpected obstacles intersecting with a mobile manipulator's trajectory. While this work focuses on avoidance, an important takeaway is the proposed use of the whole-body dynamics. 

Palejiya and Tanner \cite{Velocity_Force_Switching} implemented a switching controller that can be applied to mobile robots and uses velocity control for navigation and force control for collision recovery. Their main proposal is a strategy where a mode switch occurs when a contact force is sensed. Additionally, the paper by Berkane et al. \cite{Switching_Dimos} lays the mathematical foundations for a switching control strategy between trajectory tracking and collision avoidance.

\subsection{Contributions}
In this work we present a three mode switching NMPC control strategy for an UVMS that safely handles collisions in emergency situations, and can be applied in the emerging field of intervention tasks using AUVs. Our method offers an additional level of redundancy in cases where traditional methods offering safety guarantees, like CBFs, might fail. 

Inspiration is drawn from Nikou's high-level task MPC proposal \cite{NikouHighLevelTasks} and Chen's whole-body control for mobile manipulator collision avoidance \cite{MartialArts}, helping us formulate the objectives for the different modes used. The idea of using a switching controller for collision avoidance by Berkane \cite{Switching_Dimos} is extended to collision handling, while offering a different approach than Palejiya \cite{Velocity_Force_Switching}, by triggering a mode switch before a collision is sensed and having the robot protect itself by seeking contact with the environment before a collision occurs. This is a research direction already mentioned by Wosch \cite{CollisionHandling_MobileManipulator} in his conclusions. While some ideas and tools used in work have previously been explored, the problem we solve when combining them, to the best of the authors' knowledge, is a novel concept.

The remainder of this paper is organized as follows. Section \ref{sec:Problem_Statement}
consists of the problem statement. In Section \ref{sec:Modelling} we create the model for the robot and in Section \ref{sec:Algorithm} we formulate our control strategy. Section \ref{sec:Simulation_Results} contains results from numerical simulations. Finally, in Section \ref{sec:Conclusions_Future_Work} we draw conclusions and mention future research directions.

\section{PROBLEM STATEMENT}
\label{sec:Problem_Statement}

\subsection{Mathematical Problem Statement}
\label{subsec:MathProbStatement}

Find the optimal control inputs $\mathbf{u^*}$, and the states of the robot $\mathbf{h^*}$, with $ \{ \mathbf{u^*}, \mathbf{h^*} \} \in \mathbf{\Psi^*}$, where $\mathbf{\Psi}$ is the optimization variable vector, and $\mathbf{\Psi^*}$ is the solution to the following discrete time Optimal Control Problem (OCP).

\begin{equation}
    \begin{aligned}
        \mathbf{\Psi^*} = \ &argmin(J(\mathbf{\Psi})) \\
        s.t. \quad \quad \quad \quad &\mathbf{C^{Dyn}} = 0\\
        &\mathbf{C^{Dist}} \leq 0\\
        \mathbf{\Psi_{min}} &\leq \mathbf{\Psi} \leq \mathbf{\Psi_{max}} \\
        &\mathbf{{C_{eq}}^{Aux}} = 0 \\
    \end{aligned}
    \label{eq:OCP}
\end{equation}
The objective function given by:
\begin{equation}
    J(\mathbf{\Psi}) = 
    \begin{cases}
        J_{nom}(\mathbf{\Psi}), \quad \mathcal{F} \neq \emptyset \\
        J_{emerg}(\mathbf{\Psi}), \quad \mathcal{F} = \emptyset
    \end{cases}
\end{equation}
where $J_{nom}(\mathbf{\Psi})$ is the objective function for the vehicle mission during nominal operation, see (\ref{eq:Obj1}), while $J_{emerg}(\mathbf{\Psi})$ is an objective function used when the vehicle is on an imminent collision course, see (\ref{eq:Obj1}) (\ref{eq:Obj2}). $\mathcal{F}$ is the set of feasible solutions to the collision avoidance part of the problem according to (\ref{eq:SolutionSetAvoidance}).
\begin{equation}
    \begin{aligned}
    \mathcal{F} = \{ \mathbf{\Psi} \in \mathbb{R}^{n_{\Psi}} : \mathbf{C^{Dyn}}=0, \ &\mathbf{C^{Dist}} \leq 0, \\  &\mathbf{\Psi_{min}} \leq \mathbf{\Psi} \leq \mathbf{\Psi_{max}}  \}
    \end{aligned}
    \label{eq:SolutionSetAvoidance}
\end{equation}
where $n_{\Psi}$ is the number of variables in the optimization vector. The OCP in (\ref{eq:OCP}) features auxiliary equality constraints, $\mathbf{{C_{eq}}^{Aux}}$, that are enforced only when $\mathcal{F} = \emptyset$.


\section{MODELLING APPROACH}
\label{sec:Modelling}

\subsection{UVMS Model}
The UVMS consists of a manipulator with $N_l$ links fitted to a fully actuated AUV with 6 thrusters that is kinematically modeled as a mobile manipulator. Emphasis is given on the development of an accurate model for the highly nonlinear dynamics that govern the system. The model is extracted using Kane's method, by following Tarn et al. \cite{Tarn}, and features inertial, gravitational, hydrodynamic -added mass, buoyancy, fluid acceleration and profile drag-, control and contact forces. The vehicle state is:
\begin{equation}
       \begin{aligned}
       \mathbf{h} = 
       \begin{bmatrix}
        \mathbf{q} \\ \mathbf{\dot{q}}
       \end{bmatrix}
        = [&x \quad y \quad z \quad \phi \quad \theta \quad \psi \quad  q_{m_{1}}\ \ldots \ q_{m_{N_l}} \\
        \quad & v_x \quad v_y \quad v_z \quad \omega_x \quad \omega_y \quad \omega_z \quad \dot{q}_{m_{1}} \ \ldots \ \dot{q}_{m_{N_l}}]^{T}
        \end{aligned}
    \label{eq:SUM_State}
\end{equation}
where $[x \ y\ z]^T$, $[\phi \ \theta \ \psi]^T$, $ [v_x \  v_y \  v_z]^T $ and $[\omega_x \  \omega_y \ \omega_z]^T$ are AUV position, orientation, and linear and angular velocities respectively. $q_{m_{j}}$ and $\dot{q}_{m_{j}}$ are link angular positions and velocities. Using homogenous transfer matrices, the forward kinematics for the UVMS are given by:
\begin{equation}
    \mathbf{A_{i}^{ee}} =  \mathbf{A_{i}^0} \mathbf{A_{0}^{1}} \mathbf{A_1^2} \ldots \mathbf{A_{N_l-1}^{N_l}} \mathbf{A_{N_l}^{ee}}=
    \begin{bmatrix}
        \mathbf{R_{i}^{ee}} & \mathbf{X_{ee}} \\
        \mathbf{0_{1 \times 3}} & 1
    \end{bmatrix} \\
    \label{eq:IN2EE}
\end{equation}
where $\mathbf{X_{ee}}$ is the EE position in the inertial frame and $\mathbf{R_{i}^{ee}}$ the rotation matrix between the inertial and the EE frames.
The dynamics of the UVMS are:
\begin{equation}
    \begin{aligned}
        \mathbf{M}(\mathbf{q_m}) \mathbf{\ddot{q}} + \mathbf{C}(\mathbf{q_m}, \dot{\mathbf{q}}) + \mathbf{G}(\mathbf{q_m}) & + \mathbf{F}_{\text{ext}}(\mathbf{q}, \dot{\mathbf{q}})\\
        = &\boldsymbol{\tau}_{\text{control}} + (\mathbf{F})_{contsct}
    \end{aligned}
    \label{eq:Sum_Dynamics}
\end{equation}
with the transformation to the state-space format omitted here, see \cite{Polyzos_Thesis}. Through the differential kinematics of (\ref{eq:Jv}) we define the geometric Jacobian $\mathbf{J_v}$.
\begin{equation}
    \underset{\sim}{\mathbf{V_{ee}}} = 
    \begin{bmatrix}
        \mathbf{V_{ee}} \\
        \boldsymbol{\omega_{ee}} 
    \end{bmatrix}
    = \mathbf{J_v}(\mathbf{q})\mathbf{\dot{q}}
    \label{eq:Jv}
\end{equation}
Using (\ref{eq:F_contact}) we convert the $6 \times 1$ contact wrench from the local EE frame to the global and then into the generalized contact forces, a $(6+N_l) \times 1$ vector acting on each of the UVMS's DoF, to achieve compatibility with Kane's method. 
\begin{equation}
    (\mathbf{F})_{contact} = {\mathbf{J_v}^T}
    \begin{bmatrix}
        \mathbf{R_w} & \mathbf{0}_{3 \times 3} \\
        \mathbf{0}_{3 \times 3} & \mathbf{R_w}
    \end{bmatrix}
    \mathbf{F^w}
    \label{eq:F_contact}
\end{equation}

\subsection{Superellipsoid Body Modelling}

The proposed control strategy of Section \ref{sec:Algorithm} requires knowledge of both the distance $d_j$ and the closing speed $\dot{d}_j$ between each body and the obstacles. Any surface can be chosen to model the bodies of the UVMS in combination with any type of obstacle. The important matter is that the aforementioned quantities can be calculated quickly and online. The derivations in this work use a plane ($\mathbf{n}^T \mathbf{x} + D = 0$) as an obstacle and superellipsoids to model the surface of each of the UVMS links. The surface of a superellipsoid with parameters $\alpha, b, c, \varepsilon_1, \varepsilon_2$ is defined by (\ref{eq:superellipsoid_surface}).
\begin{equation}
    \left( \left| \frac{x}{a} \right|^{\frac{2}{\varepsilon_1}} + \left| \frac{y}{b} \right|^{\frac{2}{\varepsilon_1}} \right)^{\frac{\varepsilon_1}{\varepsilon_2}} + \left| \frac{z}{c} \right|^{\frac{2}{\varepsilon_2}} = 1 
    \label{eq:superellipsoid_surface}
\end{equation}
This approach offers analytical solutions to the distance \cite{SE_Foot_Ground} and relative velocity problems \cite{Polyzos_Thesis}.
\begin{equation}
    d_j = 
    \begin{cases}
     \| \mathbf{x_{1_j}^*} - \mathbf{x_{2_j}^*} \|, 
     & \| \mathbf{x_{2_j}^*} - \mathbf{x_{0_j}^i} \| \geq \| \mathbf{x_{1_j}^*} - \mathbf{x_{0_j}^i} \| \\

     -\| \mathbf{x_{1_j}^*} - \mathbf{x_{2_j}^*} \|, 
     & \| \mathbf{x_{2_j}^*} - \mathbf{x_{0_j}^i} \| < \| \mathbf{x_{1_j}^*} - \mathbf{x_{0_j}^i} \| 
    \end{cases} 
    \label{eq:Distance}
\end{equation}
where $\mathbf{x_{1_j}^*}$, $\mathbf{x_{2_j}^*}$ are the points of the common normal also belonging on the superellipsoid of link $j$ and the plane, respectively. Additionally,  $\mathbf{x_{0_j}}$ is the center of link $j$ that can be extracted from the kinematics.

\section{PROPOSED CONTROL ALGORITHM}
\label{sec:Algorithm}
A self-triggering switching NMPC strategy consisting of three modes is formulated as a proposed solution. Mode \MakeUppercase{\romannumeral 1} is task-oriented, Mode \MakeUppercase{\romannumeral 2} initiates contact with the obstacle, and Mode \MakeUppercase{\romannumeral 3} is used to push the vehicle away. Between the different NMPC modes, the optimization vector, $\boldsymbol{\Psi}$, defined in (\ref{eq:OptVector}), changes qualitatively.
\begin{equation}
     \boldsymbol{\Psi_{I}} = 
     \begin{bmatrix}
        \underaccent{\sim}{\mathbf{u}} \\
        \underaccent{\sim}{\mathbf{h}} \\
     \end{bmatrix}, \quad
     \boldsymbol{\Psi_{II}} = 
     \begin{bmatrix}
        \underaccent{\sim}{dt} \\
        \underaccent{\sim}{\mathbf{u}} \\
        \underaccent{\sim}{\mathbf{h}} \\
     \end{bmatrix}, \quad
     \boldsymbol{\Psi_{III}} = 
     \begin{bmatrix}
        \underaccent{\sim}{\mathbf{u}} \\
        \underaccent{\sim}{\mathbf{h}} \\
        \underaccent{\sim}{\mathbf{F^w}} 
     \end{bmatrix}, \quad
     \label{eq:OptVector}
\end{equation}

\subsection{Mode I}
\label{subsec:Mode1}
 Mode \MakeUppercase{\romannumeral 1} of the NMPC controller constitutes an OCP where the task error and the control effort of the UVMS are minimized while adhering to dynamic constraints and avoiding any obstacles present. The objective function for a prediction horizon $N_p$ is given by (\ref{eq:Obj1}).
 \begin{equation}
    J_{\MakeUppercase{\romannumeral 1}} = 
    \sum_{i=1}^{N_p} \mathbf{e}_{(k+i)}^T \mathbf{W_s} \mathbf{e}_{(k+i)} + \sum_{i=1}^{N_p} \mathbf{U}_{k}^T \mathbf{W_u} \mathbf{U}_{k}
    \label{eq:Obj1}
\end{equation}
where $\mathbf{e}_{(k+i)}$, $\mathbf{U}_{k}$ are the state error and control inputs and $\mathbf{W_s}$, $\mathbf{W_u}$ are diagonal weight matrices.

\subsection{Mode II}
\label{subsec:Mode2}

At some point, the set of feasible solutions for the optimization problem of Mode \MakeUppercase{\romannumeral 1}, $\mathcal{F}$, becomes empty, meaning that for the current prediction horizon, there is no way to avoid contact with an obstacle. The NMPC controller enters Mode \MakeUppercase{\romannumeral 2} with the goal of establishing contact with the obstacle. This is an intermediate mode that helps ensure a seamless transition into the deflection phase. The collision avoidance requirement is maintained, only permitting, and simultaneously requiring, contact between the end-effector and the obstacle at the end of the horizon. To achieve this, Mode \MakeUppercase{\romannumeral 2} uses \textbf{Algorithm \ref{alg:Mode2}}, an adaptive timestep and variable prediction horizon length algorithm. As a result, Mode \MakeUppercase{\romannumeral 2} apart from the control input vector, $\underaccent{\sim}{\mathbf{u}}$ and the dynamics state vector, $\underaccent{\sim}{\mathbf{h}}$ also optimizes the timestep, $dt$. The objective function is given by (\ref{eq:Obj2}).
\begin{equation}
    J_{\MakeUppercase{\romannumeral 2}} = \frac{1}{{d_{0}}^{k|k+N_p} + \varepsilon} + w\| {\mathbf{V_{ee}}}^{k|k+N_p} \|
    \label{eq:Obj2}
\end{equation}
Here ${d_{0}}^{k|k+N_p}$ is the distance of the UVMS's body from the obstacle at the end of the horizon, as calculated at timestep $k$, w is a weight between objectives and $\varepsilon$ is a small quantity.
\begin{algorithm}[htbp]
\caption{Mode \MakeUppercase{\romannumeral 2} Variable Horizon Length Algorithm}
\label{alg:Mode2}
    \begin{algorithmic}[1]
        \STATE Problem initialization from Mode \MakeUppercase{\romannumeral 1}
        \WHILE{in Mode \MakeUppercase{\romannumeral 2}}
            \STATE Solve OCP with prediction horizon $N_p^{(k)}$ to get $dt^*$
            \STATE Take an NMPC step with $dt^*$
            \STATE Calculate remaining expected time till contact: \\ $t_{exp} = (N_p^{(k)}-1)dt^*$ 
            \IF{$N_p^{(k)} > 1$}
                \STATE Calculate new prediction horizon length: \\ $N_p^{(k+1)} = ceil(t_{exp}/dt_{nom})$
                \STATE $Mode^{(k+1)} \leftarrow \MakeUppercase{\romannumeral 2}$
            \ELSE 
                \STATE Use nominal prediction horizon length: \\$N_p^{(k+1)} \leftarrow N_p$
                \STATE $Mode^{(k+1)} \leftarrow \MakeUppercase{\romannumeral 3}$
            \ENDIF
            \STATE $k \leftarrow k + 1$
        \ENDWHILE
        \STATE \textbf{return} result
    \end{algorithmic}
\end{algorithm}

\subsection{Mode III}
\label{subsec:Mode3}

With the end-effector now in contact with the obstacle, the UVMS uses the manipulator to exert the necessary force to avoid the collision. The OCP requires that the solution conforms with the dynamics constraints while also maintaining contact with the obstacle. The target is the minimization of the relative velocity between the AUV body and the obstacle. The optimization variable vector now contains the contact forces and torques acting on the UVMS expressed in the local obstacle coordinate frame, $\underaccent{\sim}{\mathbf{F^w}}$. The objective function is:
\begin{equation}
    J_{\MakeUppercase{\romannumeral 3}} = \sum_{i = 1}^{N_p}w_i {\dot{d}_0}^{k|k+i}
    \label{eq:Obj3}
\end{equation}
where $w_i$ are weights given by:
\begin{equation}
    w_i =
    \begin{cases}
        1 + \frac{N_p-i}{N_p-1}, \quad \quad &{\dot{d}_0}^{k|k+i}<0 \\
        0, &{\dot{d}_0}^{k|k+i} \geq 0
    \end{cases}
\end{equation}

\textbf{Algorithm \ref{alg:Mode3}} is developed to aid with the initialization of an appropriate contact strategy implemented by the manipulator. In order for the vehicle to push itself away from the obstacle it is safe to assume that the (negative) normal force should be minimized, while not overstressing the actuators of the manipulator from the requested torque $\mathbf{T_m}$.
\begin{equation}
    \begin{aligned}
    \min \quad \quad&F^w_x \\
    \text{subject to} \quad \quad &(\mathbf{T_{m}})_{min} \leq \mathbf{T_{m}} \leq (\mathbf{T_{m}})_{max},
    \end{aligned}
    \label{eq:Fc_Init_OptProblem}
\end{equation}
At the same time we require that the EE maintains contact with the obstacle, and that no kinematic constraints are violated. Due to the momentum of the vehicle the manipulator needs to change its posture. This is initialized from the optimization problem (\ref{eq:Wall_InvKin}) which is basically the inverse kinematics of the redundant UVMS.
\begin{equation}
    \begin{aligned}
    \min \quad & (\mathbf{q_m} - \mathbf{q_m}^{k|k+i-1})^T(\mathbf{q_m} - \mathbf{q_m}^{k|k+i-1}) \\
    \text{subject to} \quad & {C_{eq}}^{EE} =  \frac{\mathbf{n}^T\mathbf{X_{ee}([q_{AUV};q_m])}+D}{\|\mathbf{n}\|} = 0 \\
                             & \mathbf{q_m}_{min} \leq \mathbf{q_m} \leq \mathbf{q_m}_{max}
    \end{aligned}
    \label{eq:Wall_InvKin}
\end{equation}
In (\ref{eq:Wall_InvKin}), the optimization vector is $\mathbf{q_m} \equiv \mathbf{q_m}^{k|k+i}$.
\begin{algorithm}[htbp]
\caption{Mode \MakeUppercase{\romannumeral 3} Contact Force Initialization}
\label{alg:Mode3}
\begin{algorithmic}[1]
    \STATE Problem initialization from Mode \MakeUppercase{\romannumeral 2}
    \FOR{each step $k+i$, where $i = 1,\dots,N_p$}
        \STATE Solve (\ref{eq:Fc_Init_OptProblem}) for contact force initialization
        \STATE Calculate a UVMS state guess, $\mathbf{h^{k|k+1}}$, using the state space model of the dynamics (\ref{eq:Sum_Dynamics}), based on:
        \begin{itemize}
            \item the current state,
            \item the contact force initialization, and
            \item the last known control inputs
        \end{itemize}
        \STATE Keep the AUV states, $\mathbf{q_{AUV}}$, from the guess $\mathbf{h^{k|k+1}}$
        \STATE Solve the inverse kinematics (\ref{eq:Wall_InvKin}) to find manipulator posture $\mathbf{q_m}$ using the AUV state guess 
        \STATE Assume $\dot{\mathbf{q}}_m \approx 0$
        \STATE Combine to form the UVMS state initialization, $\mathbf{\Psi}^k_{\text{init}}$
    \ENDFOR
    \STATE \textbf{return} $\mathbf{\Psi}^k_{\text{init}}$
\end{algorithmic}
\end{algorithm}
When the body of the UVMS is no longer approaching the obstacle, the manipulator has served its purpose in protecting the AUV from damage through the use of Modes \MakeUppercase{\romannumeral 2} \& \MakeUppercase{\romannumeral 3} and the vehicle can return to its initial task in Mode \MakeUppercase{\romannumeral 1}.

\subsection{Mode Switching \& Initializations}
\begin{figure*}[!t]
    \centering
    \includegraphics[width=1\textwidth]{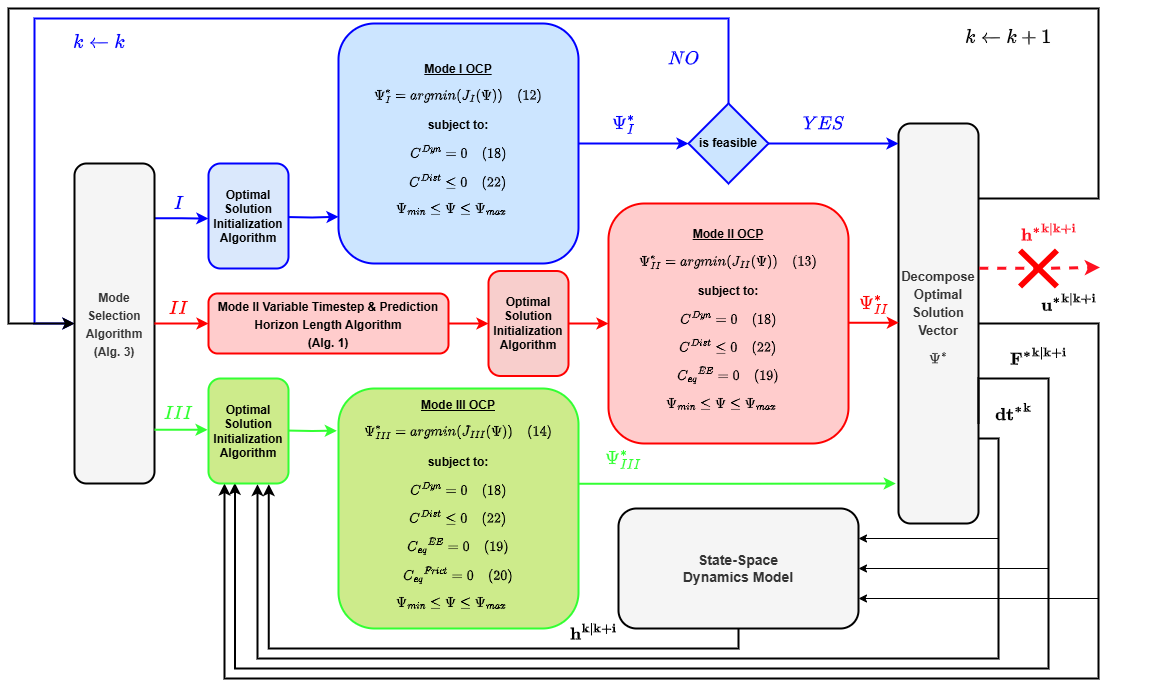}
    \captionsetup{font=small, justification=centering}
    \caption{Switching Nonlinear Model Predictive Control Algorithm.}
    \label{fig:Algorithm}
\end{figure*}

After formulating the three different controller modes it is important to select which on is applied at every timestep $k$. This is done through \textbf{Algorithm \ref{alg:ModeSelection}}.
\begin{algorithm}[b]
\caption{Mode Selection – Mode Switching Conditions}
    \begin{algorithmic}[1]
        \IF{$\mathcal{F}^{(1)} \neq \emptyset$}
            \STATE $Mode^{(1)} \leftarrow \MakeUppercase{\romannumeral 1}$
        \ELSE
            \STATE $Mode^{(1)} \leftarrow \MakeUppercase{\romannumeral 2}$
        \ENDIF
        \STATE $k \leftarrow 2$
        \WHILE{running the control algorithm}
            \IF{$Mode^{(k-1)} = \MakeUppercase{\romannumeral 1}$ \AND $\mathcal{F}^{(k)} = \emptyset$}
                \STATE $Mode^{(k)} \leftarrow \MakeUppercase{\romannumeral 2}$
            \ELSIF{$Mode^{(k-1)} = \MakeUppercase{\romannumeral 2}$ \AND ${N_p}^{(k-1)} = 1$}
                \STATE $Mode^{(k)} \leftarrow \MakeUppercase{\romannumeral 3}$
            \ELSIF{$Mode^{(k-1)} = \MakeUppercase{\romannumeral 3}$ \AND ${\dot{d}_0}^{(k-1)} > 0$}
                \STATE $Mode^{(k)} \leftarrow \MakeUppercase{\romannumeral 1}$
            \ELSE
                \STATE $Mode^{(k)} \leftarrow Mode^{(k-1)}$
            \ENDIF
            \STATE $k \leftarrow k + 1$
        \ENDWHILE
        \STATE \textbf{return} $Mode^{(k)}$
    \end{algorithmic}
    \label{alg:ModeSelection}
\end{algorithm}
Mode chattering is not an issue for this switching algorithm, in part due to the linear fashion of mode switching (\MakeUppercase{\romannumeral 1} $\rightarrow$ \MakeUppercase{\romannumeral 2} $\rightarrow$ \MakeUppercase{\romannumeral 3} $\rightarrow$ \MakeUppercase{\romannumeral 1}), but mainly due to the fact that Mode \MakeUppercase{\romannumeral 1} should always be feasible after Mode \MakeUppercase{\romannumeral 3} has terminated. This is based on the assumptions of guaranteed Mode \MakeUppercase{\romannumeral 1} feasibility when not on a collision course, and the Mode \MakeUppercase{\romannumeral 3} termination condition. However, this control strategy is sensitive to initializations, especially when mode switching occurs. The standard MPC receding horizon procedure is applied in conjunction with \textbf{Algorithms \ref{alg:Mode2} \& \ref{alg:Mode3}}. \textbf{Algorithm \ref{alg:Mode3}} is a critical part of the optimal solution initialization algorithm (Fig. \ref{fig:Algorithm}) that is omitted here, but further intricacies that enhance the MPC performance are discussed in \cite{Polyzos_Thesis}.

\subsection{Constraints}

Next we extract the equations of the OCP constraints for the different modes. Equation (\ref{eq:DynConstraints}) describes the dynamics constraints for the full prediction horizon.
\begin{equation}
    \mathbf{C^{Dyn}} = 
    \begin{bmatrix}
        \underaccent{\sim}{\mathbf{h}}^k_{1:NoS} - \mathbf{h}^{k|k+1} \\ 
        \underaccent{\sim}{\mathbf{h}}^k_{(NoS+1):2NoS} - \mathbf{h}^{k|k+2} \\ 
        \vdots \\
        \underaccent{\sim}{\mathbf{h}}^k_{(NoS(N_p-1)+1):NoS*N_p} - \mathbf{h}^{k|k+N_p} \\ 
    \end{bmatrix}
    \label{eq:DynConstraints}
\end{equation}
Optimization variables are denoted with $\underaccent{\sim}{\mathbf{h}}$ while states that satisfy the dynamics (\ref{eq:Sum_Dynamics}) are denoted as $\mathbf{h}$ and $NoS$ is the number of states.

The contact constraint between the manipulator's end-effector and the obstacle is given by:
\begin{equation}
    {C_{eq}}^{EE}_i =  \frac{\mathbf{n}^T\mathbf{X_{ee}(\underaccent{\sim}{q}^{k|k+i})}+D}{\|\mathbf{n}\|}
    ,  
    i = \begin{cases}
        N_p & \text{Mode II} \\
        1,\ldots,N_p & \text{Mode III} \\
    \end{cases}
    \label{eq:EE_Wall_Dist_Constraint}
\end{equation}
Friction is implemented with the constraints:
\begin{equation}
    {C_{eq}}^{Frict}_i =  
    \begin{bmatrix}
        \mu |F^w_{x_i}| - \| \mathbf{F^w_{tan_i}}\| \\
        \mathbf{F^w_{tan_i}} \cdot \mathbf{e^{V_{ee}}_i} + \| \mathbf{F^w_{tan_i}} \|*\|\mathbf{e^{V_{ee}}_i}\|
    \end{bmatrix}
    \label{eq:Friction_Constraint}
\end{equation}
where:
\begin{equation}
    \mathbf{F^w_{tan}} = [0\quad F^w_y \quad F^w_z]^T
    \label{eq:F_wall_tan}
\end{equation}
$\mu$ is the coefficient of friction and $\mathbf{e^{V_{ee}}}$ is the unit vector projection of the EE velocity on the plane.

Safe distance between each UVMS body and the obstacle, for the duration of the prediction horizon, is achieved using the inequality constraints in (\ref{eq:DistConstraints}).
\begin{equation}
    {C^{Dist}_{j,i}} = -d_j^{k|k+i} \leq 0
    \label{eq:DistConstraints}
\end{equation}

\subsection{Final Algorithm}
The proposed control strategy is summarized in the flowchart of Fig. \ref{fig:Algorithm}. Besides the final formulations of the constrained OCPs and the algorithms previously presented, our approach incorporates another intricacy. Since a numerical solution of the OCPs uses a tolerance for constraint violations, due to the way that the dynamic constraints are formulated in (\ref{eq:DynConstraints}) those can potentially propagate throughout the prediction horizon. The performance of the algorithm can be improved by recalculating the state trajectory $\mathbf{h}^{k|k+i}$ from the dynamic model using the optimal control inputs, $\mathbf{u}^{*k|k+i}$, contact force, $(\mathbf{F^{w^*}})^{k|k+i}$ and timestep, $dt^{*k}$, before feeding them into the initialization algorithm for the next timestep \textit{k}.

\section{SIMULATION \& RESULTS}
\label{sec:Simulation_Results}

A simulation environment was created in MATLAB. The kinematic and dynamic model of the UVMS was developed by realizing \cite{Tarn} using the symbolic math toolbox. The control strategy is implemented and the robot's telemetry is extracted along with animations \cite{Polyzos_Thesis}.

\begin{figure}[!b]
    \centering
    \includegraphics[width=0.95\columnwidth]{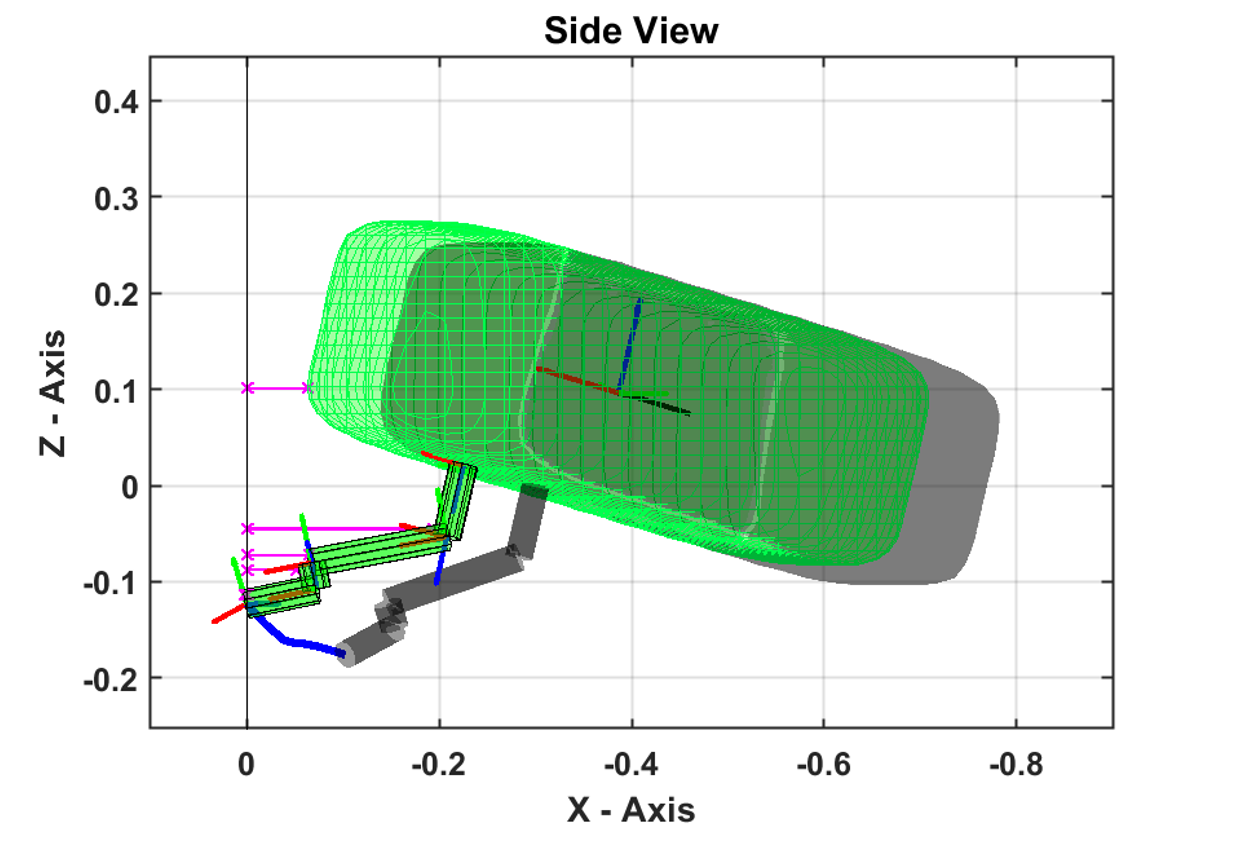}
    \captionsetup{font=small, justification=centering}
    \caption{UVMS in contact with obstacle: \\Mode III - NMPC Step 9 - Case 2B from \cite{Polyzos_Thesis}.}
    \label{fig:UVMS}
\end{figure}
\begin{figure}[!b]
    \centering
    \includegraphics[width=\columnwidth]{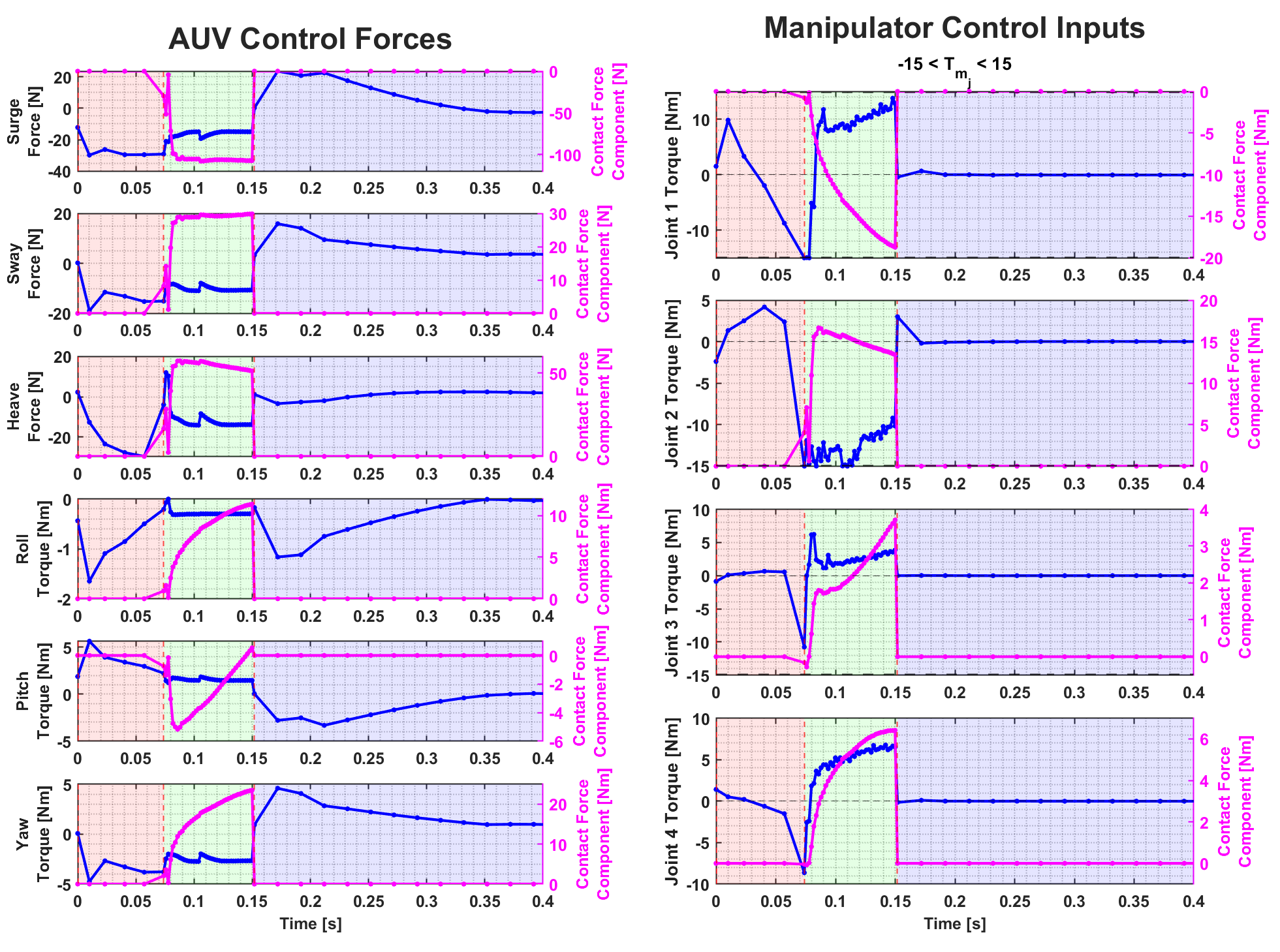}
    \captionsetup{font=small, justification=centering}
    \caption{UVMS Control Forces - Case 4B from \cite{Polyzos_Thesis}.}
    \label{fig:Control_Inputs}
\end{figure}
\begin{figure*}[!t]
    \makebox[\textwidth][c]{%
        \includegraphics[width=0.98\textwidth]{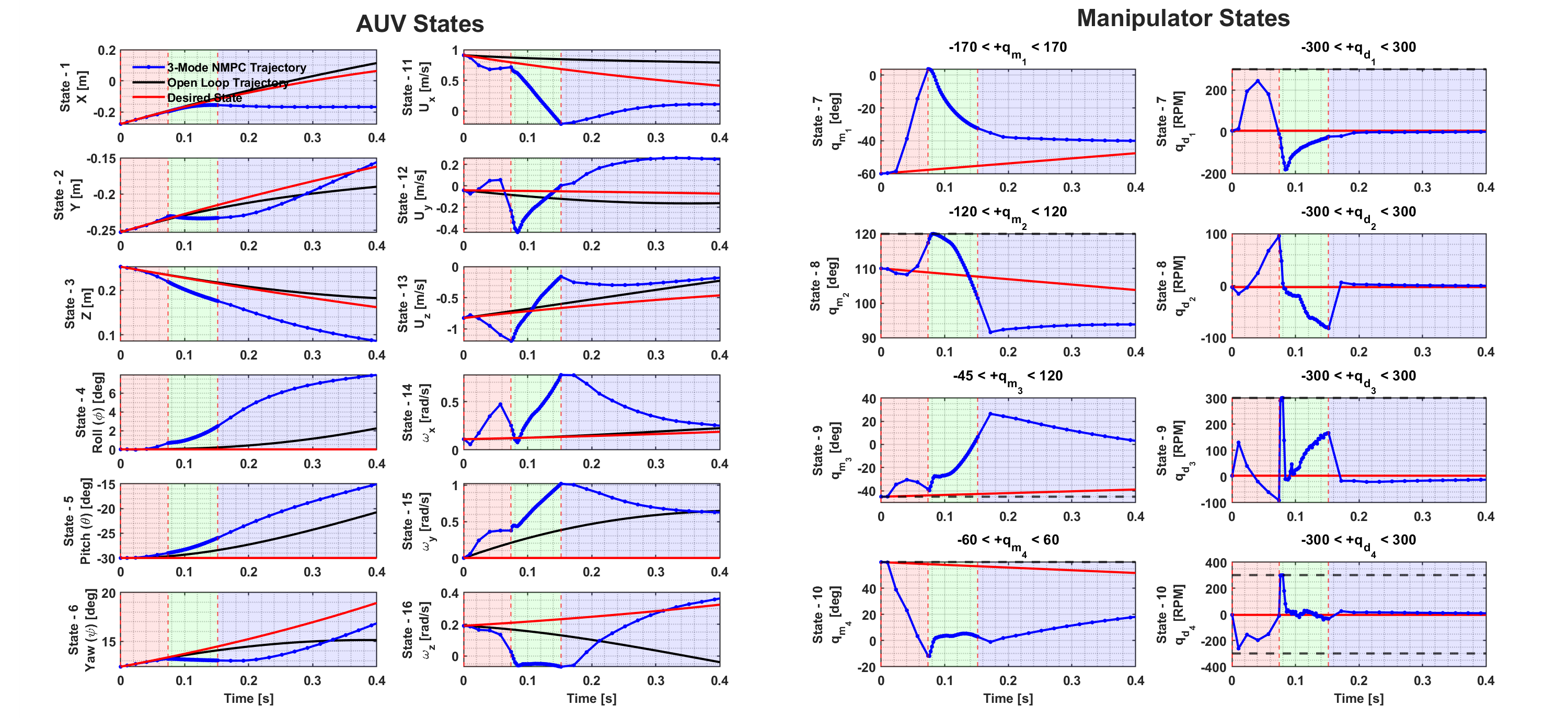}
    }
    \captionsetup{font=small, justification=centering}
    \caption{UVMS States Telemetry - Case 4B from \cite{Polyzos_Thesis}}
    \label{fig:Sum_States}
\end{figure*}

Simulations where conducted for various initial conditions, vehicle missions and environments such as the presence of ocean currents, different manipulator configurations and changes to the parameters of the control algorithm. Additionally the scenario of full thruster malfunction was simulated and the manipulator is seen in action in Fig. \ref{fig:UVMS}. The proposed strategy successfully handles these cases, now presenting control input (Fig. \ref{fig:Control_Inputs}) and state response (Fig. \ref{fig:Sum_States}) telemetry from Case 4B in \cite{Polyzos_Thesis}. Here the vehicle is in a dire situation and Mode \MakeUppercase{\romannumeral 2} is immediately triggered. The transfer of the contact force to the vehicle through the torques of the manipulator joints that leads to the reduction of speed and momentum for the vehicle needs to be emphasized. Note that the manipulator can exert a stopping significantly larger than the thrusters. Telemetry and animations are available at \linebreak DOI: 10.5281/zenodo.18357280.

\section{CONCLUSIONS \& FUTURE WORK}
\label{sec:Conclusions_Future_Work}
We demonstrate that voluntary contact initiation, on better terms, can be used to mitigate escalation of a situation. The strategy presented can handle a significant subset of scenarios where an UVMS is on an imminent collision course with an obstacle. This allows for an extension of the operational envelope of the UVMS without increasing the risk of sustaining damage. The controller architecture developed is non-invasive during nominal robot operation and can be attached to any other existing controller for an UVMS acting as a redundancy in case of emergency. This switching control architecture is not limited to an UVMS but can also be implemented on a mobile manipulator platforms and humanoid robots. 

Future research directions include reformulating Mode \MakeUppercase{\romannumeral 3} in order for the algorithm to maintain operability till the moment of impact even in extreme cases where the body's collision remains imminent, further increasing the operational envelope. Additional improvements include substituting the heuristic optimal solution initialization algorithm, with a Neural Network trained via Reinforcement Learning on the UVMS model, that generates feasible initial guesses. Finally, it is suggested replacing MATLAB's fmincon with a custom NMPC solver that balances feasibility and optimality.


\end{document}